\documentclass[10pt, a4paper]{article}
\usepackage{lrec2022} 
\usepackage{multibib}
\newcites{languageresource}{Language Resources}
\usepackage{graphicx}
\usepackage{tabularx}
\usepackage{soul}
\usepackage{footnote}

\usepackage{titlesec}
\titleformat{\section}{\normalfont\large\bfseries\center}{\thesection.}{1em}{}
\titleformat{\subsection}{\normalfont\SmallTitleFont\bfseries\raggedright}{\thesubsection.}{1em}{}
\titleformat{\subsubsection}{\normalfont\normalsize\bfseries\raggedright}{\thesubsubsection.}{1em}{}
\renewcommand\thesection{\arabic{section}}
\renewcommand\thesubsection{\thesection.\arabic{subsection}}
\renewcommand\thesubsubsection{\thesubsection.\arabic{subsubsection}}

\usepackage{epstopdf}
\usepackage[utf8]{inputenc}

\usepackage{hyperref}
\usepackage{xstring}

\usepackage{color}

\title{ASL-Homework-RGBD Dataset: An annotated dataset of 45 fluent and non-fluent signers performing American Sign Language homeworks}

\name{\begin{small}Saad Hassan\textsuperscript{1}, Matthew Seita\textsuperscript{1}, Larwan Berke\textsuperscript{1}, Yingli Tian\textsuperscript{2}, Elaine Gale\textsuperscript{3} , Sooyeon Lee\textsuperscript{1}, Matt Huenerfauth\textsuperscript{1}\end{small}} 

\address{\textsuperscript{1} Rochester Institute of Technology, Rochester, NY 14623 USA \\ \textsuperscript{2} The City College of New York, New York, NY 10031 US, \\
         \textsuperscript{3} Hunter College, City University of New York, 695 Park Ave, New York, NY 10065\\
         sh2513@rit.edu , mss4296@rit.edu, lwb2627@rit.edu,  
         \\ ytian@ccny.cuny.edu , egale@hunter.cuny.edu , slics@rit.edu , matt.huenerfauth@rit.edu \\
         }

\abstract{
We are releasing a dataset containing videos of both fluent and non-fluent signers using  American Sign Language (ASL), which were collected using a Kinect v2 sensor. This dataset was collected as a part of a project to develop and evaluate computer vision algorithms to support new technologies for automatic detection of ASL fluency attributes. A total of 45 fluent and non-fluent participants were asked to perform signing homework assignments that are similar to the assignments used in introductory or intermediate level ASL courses. The data is annotated to identify several aspects of signing including grammatical features and non-manual markers. Sign language recognition is currently very data-driven and this dataset can support the design of recognition technologies, especially technologies that can benefit ASL learners. This dataset might also be interesting to ASL education researchers who want to contrast fluent and non-fluent signing. 
 \\ \newline  }

\begin{document}

\maketitleabstract

\section{Background and Related Work}

Advancements in deep learning and sensor technologies as well as research on computer vision techniques have enabled the development of sign language recognition systems \cite{rastgoo2021sign}. While the accuracy of sign language recognition technologies have improved, there are still some challenges that need to be resolved. Modern machine learning approaches to sign-recognition based on neural networks are largely data-driven; however, current publicly released datasets of sign languages are still several orders smaller in magnitude compared to datasets of other spoken languages used to train automatic speech recognition systems. 

While summarizing the challenges facing the sign-recognition field, a recent paper identified 4 dimensions on which to classify datasets: size, continuous real-world signing, the inclusion of native signers, and signer variety \cite{sign_language_recognition_generation_translation}. Since datasets of isolated signs can only support very specific use-cases, e.g. sign language dictionaries, it is therefore important to collect continuous signing datasets (natural conversational data or at least longer utterances) from a diverse set of signers to support useful real-world applications \cite{albanie2021bbc}.

Existing datasets of ASL usually consist of videos of people performing continuous signs \cite{sign_language_recognition_generation_translation,albanie2021bbc}, e.g. How2Sign \cite{duarte2021how2sign}, NCSGLR \cite{databases2007volumes}, ASLG-PC$_{12}$ \cite{othman2012english}, CopyCat \cite{zafrulla2010novel}, RWTH-BOSTON-400 and RWTH-BOSTON-104 \cite{RWTH,dreuw:interspeech2007}. There are some datasets of isolated signs, e.g. ASL-LEX-2.0 \cite{asl-lex2.0}, WLASL \cite{li2020word}, ASL-100-RGBD \cite{hassan-etal-2020-isolated}, MSASL \cite{vaezijoze2019ms-asl}, ASL-LEX \cite{caselli2017asl}, ASLLVD \cite{athitsos2008american}, Purdue RVL-SLL \cite{martinez2002purdue}, etc., and fingerspelling as well, e.g. ChicagoFSWild+ \cite{shi2019fingerspelling} and ChicagoFSWild \cite{shi2018american} . Table \ref{table1} 
describes some of these datasets in greater detail \footnote{Table 2 in \cite{albanie2021bbc} enlists summary statistics of other sign language datasets.}.

\begin{table*}[]
\scriptsize
\begin{tabular}{|l|l|l|l|l|l|l|l|l|l|}
\hline
              Dataset  & Year & Type                                                          & Multi & Pose & Depth  & Samples     & Signers & F/nF & Publication \\ \hline

ASLLRP      & 2022 & Isolated                                                &                  &      &       &       23452         & 33    &  F   & \cite{neidle:22037:sign-lang:lrec}          \\ \hline
ASL-LEX 2.0     & 2021 & Isolated                                                &                  &      &       &       2723         & 1    &  F   & \cite{asl-lex2.0}            \\ \hline
WLASL           & 2020 & Isolated                                                &                  &      &       &        21083        & 119  &  F  &  \cite{li2020word}            \\ \hline
ASL-100-RGBD    & 2020 & Isolated                                                &                  &      & Yes   &       4150         & 22  &   F  & \cite{hassan-etal-2020-isolated}             \\ \hline
How2Sign        & 2020 & Continuous                                            & Yes              & Yes  & Yes   &        35000        & 11  &  F   & \cite{duarte2021how2sign}            \\ \hline
MS-ASL          & 2019 & Isolated                                                &                  &      &       &        25000        & 200  & F   & \cite{vaezijoze2019ms-asl}            \\ \hline
ChicagoFSWild+  & 2019 & Fingerspelling                                                           &                  &      &              & 55232          & 260  &  F  &           \cite{shi2019fingerspelling}  \\ \hline
ChicagoFSWild   & 2018 & Fingerspelling                                                              &                  &      &       &        7304    &  200  & F  &   \cite{shi2018american}          \\ \hline
ASL-LEX         & 2016 & Isolated                                                &                  &      &       &        1000         & 1       & F &    \cite{caselli2017asl}         \\ \hline
NCSGLR          & 2012 & Continuous & Yes              &      &       &        11,854          & 8      & F  &          \cite{databases2007volumes}   \\ \hline
CopyCat         & 2010 & Continuous &                  &      &       &    420          & 5     & nF  &    \cite{zafrulla2010novel}         \\ \hline
ASLLVD          & 2008 & Isolated                                                & Yes              & Yes  &       &        9748        & 6      & F &   \cite{athitsos2008american}          \\ \hline
RWTH-BOSTON-400 & 2008 & Continuous  & Yes              &      &       &        -           & 4      & F &        \cite{RWTH}     \\ \hline
RWTH-BOSTON-104 & 2007 &Continuous & Yes              &      &       &        -           & 3      & F &   \cite{dreuw:interspeech2007}          \\ \hline
Purdue RVL-SLL  & 2002 & Isolated                                                &                  &      &       &        -      & 14  &  F   &  \cite{martinez2002purdue}           \\ \hline
\textbf{ASL-Homework-RGBD}   & \textbf{2022} & \textbf{Continuous}                                                &                  &      & \textbf{Yes}       &         \textbf{935}       & \textbf{45}  &  \textbf{Both}   &            \\ \hline
\end{tabular}
\normalsize
\label{table1}
\caption{Examples of published ASL datasets, with the year of release and the type of signing it contains (Isolated, Continuous, or Fingerspelling).  The table indicates whether multiple camera views (e.g., front and side) were included (Multi), whether 3D human skeleton information is included (Pose), whether RGBD depth information is included (Depth), the number of videos (Samples), the number of people in the dataset (Signers), whether the signers were fluent, non-fluent, or both (F/nF), and a citation (Publication). The last row describes the ASL-Homework-RGBD dataset shared with this paper. The non-fluent (``nF") participants in the COPYCAT dataset included Deaf children with developing ASL skills.}
\end{table*}

Data collection methodologies and apparatuses as well as the motivations behind data collection effort determine what the final publicly released datasets look like.  Datasets have been collected to support sign recognition efforts (training and benchmark testing sets), generate signing avatars, and design systems for learning different sign languages. For example, motion capture datasets that make use of sensors attached to signers are often curated to generate signing avatars \cite{lu2010collecting,heloir2005captured,berret2016collecting}. Datasets also vary on the demographic profiles of the signers and geographic regions in which they are collected. The demographic profiles can include paid professional interpreters on live TV \cite{forster2014extensions,koller2017re} and research assistants hired to record \cite{martinez2002purdue,zahedi2006continuous}, ASL students, or Deaf signers, etc. Datasets can be collected in controlled laboratory settings or collected using scrapping online video libraries and sites, e.g. YouTube \cite{joze2018ms}.

Another key aspect of the publicly released sign language datasets is their annotations. Annotations can be in the form of the closest English label or gloss, which are linguistic notations representing each sign component, or just translated text. Annotations can also demarcate signs in different manners, e.g. start and end of each handshape, sign, or a phrase/sentence. Specialized analysis software resources may also be employed, e.g ELAN \cite{elan}, SignStream  \cite{neidle2018new,neidle2020user}, VIA  \cite{dutta2019via}, iLex \cite{hanke2002ilex}, or Anvil  \cite{kipp_2017}. In collection of some of the datasets, researchers also engaged Deaf annotators for a manual-sign annotation-verification step at the end \cite{albanie2021bbc}.

With this paper, we are releasing an annotated dataset of continuous ASL signing from 45 signers. A unique contribution of our new dataset is that it includes recordings of both fluent and non-fluent ASL signers, who are engaged in the same set of homework-style expressive signing tasks. In addition, the annotation of our dataset not only includes gloss labels and annotation of syntactic non-manual expressions, but it also includes labels as to whether specific errors have occurred in the signing, e.g., when a non-fluent signer may have omitted a linguistically required non-manual expression. Given these characteristics, our dataset may be useful for research on detection of production errors in ASL signing, e.g., in the context of educational systems, and this data may also be of interest to educational or linguistics researchers, who wish to compare ASL production among signers of various levels of fluency.

We describe the context and motivation of our work in section 2. We then describe the dataset in detail including the apparatus used, data collection methods, participant recruitment, and post-processing of the data. Finally, in section 4, we conclude with the insights we learned and some of the limitations of the dataset.

\section{Context of Data Collection and Release}

This is a novel dataset that has been collected as a part of a collaborative project between Rochester Institute of Technology, The City College of New York, and Hunter College \cite{vahdani2021recognizing,10.1145/3046788,7532886}. A previous dataset of isolated ASL signs for the educational tool was released at LREC 2020 \cite{hassan-etal-2020-isolated}.

This paper describes a video-recording corpus of students (and fluent signers) performing ASL phrases and sentences, as a part of homework assignments. This new dataset was collected to support the design of technologies to fundamentally advance partial-recognition of some aspects of ASL. For example, identifying linguistic and performance attributes of ASL without necessarily determining the entire sequence of signs, or automatically determining if a performance is fluent or contains errors made by ASL students. This research effort was aimed at enabling future computer-vision technologies to support educational tools to assist ASL learners in achieving fluency, with an automatic system providing feedback on their signing. The ASL-Homework-RGBD dataset is available to authorized users of the Databrary platform\footnote{https://nyu.databrary.org/volume/1249}. RGBD in the dataset title refers to red-green-blue color video and depth information, provided by a color and depth camera, such as the Kinect.

\section{ASL-Homework-RGBD Dataset}

\subsection{Participants and Recruitment}

We recruited 45 ASL signers to be recorded in this IRB-approved data collection effort, using electronic and paper advertisements across the Rochester Institute of Technology and National Technical Institute for the Deaf campus.  Our participants consisted of 24 fluent signers and 21 non-fluent students.

Our fluent participants included 17 men and 7 women aged 20 to 51 (mean=25.08, median=23 years, SD=6.65). 5 of the participants self-described as Hard-of-hearing while the rest 19 self-described as Deaf/deaf. To recruit fluent ASL signers, we used the following screening questions:  Did you use ASL at home growing up, or did you attend a school as a very young child where you used ASL? 
 
Our non-fluent participants included 6 men and 15 women aged 18 to 49 (mean=23.19, median=21 years, SD=7.65). 4 of the participants self-described as Hard-of-hearing while the rest 17 self-described as hearing. To recruit non-fluent ASL signers (students who were learning the language), we used the following screening questions: Are you currently taking an introductory or intermediate course in American Sign Language, or have you completed an introductory or intermediate ASL course in the past five years?  

\subsection{Data Collection Procedure and Apparatus} 

Each participant was assigned a codename starting with ``P" if they were a not-fluent signer, e.g. P01, or ``F" if they were a fluent signer, e.g. F13. A paper copy of a consent form was shared with the participants which they signed. They then filled out a short demographic questionnaire. 

Participants were told: \textit{You will work on a “homework” style assignment, from an American Sign Language class, where you will need to make a video of yourself signing.}  We shared a paper copy of the homework-assignment prompt that they would be attempting during the session.  (Details of these prompts appear below.)  Some participants, especially fluent signers, had time to complete multiple homework assignments during a single one-hour recording session visit, and other participants returned to the laboratory on multiple days for additional sessions, to complete additional assignments.  The camera was 1.5m from the signer, and there was a researcher in the room with the participant.  For hearing students learning ASL, this was a hearing researcher, and for Deaf fluent ASL signers, this was a Deaf ASL-signer researcher. Participants were given \$40 (U.S. dollars) compensation for participating in each one-hour recording session.  

When considering the prompt and preparing what they would like to sign in ASL, a hard copy of an ASL-English dictionary and some other ASL reference books were made available to participants.  They were encouraged to write a script first and practice so that they could produce their signing for each video without looking at their paper.   The researcher was told to make sure that there was at least 30 minutes available to do the recording, and thus, if a participant was taking over 20 minutes to prepare for their signing, the researcher needed to encourage them to finish up their preparation soon. 

The researcher then made sure that the Kinect v2 camera system was working properly, that it was recording at approximately 30 frames per second (FPS), and that the system was detecting a ``skeleton" pose of the participant. Each video recording was assigned a codename in the format ParticipantID-UtteranceNumber, e.g., for non-fluent participant 1 and utterance 1 the name assigned was P01U01. (In this dataset, each individual video that was produced is referred to as an utterance.)  

Participants were discouraged from signing any introductory information at the beginning of their video, e.g., ``Phrase Number 1."   The researcher switched off recording as soon as the participant finished. Participants were strongly encouraged to use a standard starting and ending position (hands on their lap). If participants attempted a phrase multiple times, only the last video was kept. 

\subsection{Description of Prompts}

As stimuli prompts for signers, a series of homework assignments were created, to align with concepts traditionally taught in a second-semester ASL course at the university level.  Some of these prompts asked the signer to produce a sequence of 1-2 sentence videos, and other prompts asked the signer to produce a longer multi-sentence video.  In total, there were 6 homework prompts, with each focusing on a set of grammatical concepts, as described below. The homework prompts are also shared with the dataset.

\subsubsection{Homework 1: WH Questions and YN Questions} This assignment consisted of 10 short prompts, each of which required the signer to produce a single question. Participants were asked to use non-manual signals (e.g., facial expressions and head movements) correctly as they produced these WH and Yes-No questions. The English text descriptions (of what to ask about) encouraged the signer to produce questions that, at times, contained some fingerspelling, numbers, or pointing to locations in the signing space to refer to people. 

\subsubsection{Homework 2: Your Autobiography} Participants were asked to produce a multi-sentence ASL passage about themselves. Some key information that they were asked to include were their name, whether they are deaf or hearing, what languages they know, their high school and college education, some activities that they were part of in high school and college that they liked or disliked (using a contrastive structure), etc.

\subsubsection{Homework 3: Describing Pets} This prompt consisted of two open questions, each of which encouraged the signer to produce a short multi-sentence passage. In the first question, they were asked to compare and contrast two pets that they have or wished they had. In the second question, they were required to invent and ask 4 questions related to pets (directing the question to the camera). 

\subsubsection{Homework 4: Your Home} This assignment asked signers to produce one multi-sentence video to discuss where they live, the type of home they live in, their neighborhood, where they work or go to school, their commute to work or school, and who they live with.  

\subsubsection{Homework 5: Pronouns and Possessives} This assignment consisted of 12 short prompts, each of which consisted of two English sentences.  Participants were asked to produce a short video for each, to convey the meaning in ASL.  The sentences were specifically designed to include many personal pronouns (e.g., you, me, him) and possessive adjectives (e.g, your, my, his). 

\subsubsection{Homework 6: Conditional Sentences and Rhetorical Questions} This assignment consisted of 12 short prompts in written English that students had to translate into ASL, to produce a short video for each prompt. The sentences were designed so that the ASL signing would likely require the signer to produce conditional phrases or rhetorical questions.

\subsection{Description of Annotation Process}

After each recording session, the video files were converted to the MOV format for analysis within the ELAN analysis tool \cite{elan} and for distribution in our dataset. Our team of annotators included both ASL interpreters (who had completed a semester-long university course in ASL linguistics) and DHH individuals with native-level fluency in ASL (who received training on the specific linguistic properties being labeled).  Our annotation and analysis process consisted of a four-pass process: First, one of the ASL interpreters on the project analyzed each video. Second, it was cross-checked by another ASL interpreter on the team for accuracy.  Third, it was checked by a DHH researcher on our team with native-level ASL fluency, and finally, it was checked by a faculty member with expertise in ASL linguistics.

There were 6 different groups of annotation tiers, and annotators were encouraged to go from the simplest one and gradually move to more complex tiers.  We describe each group of annotation tiers in this section in a similar manner:  

The first tier, \textit{Signing Happening}, was used to just identify the times when any signing is happening. The next tier was \textit{Timing of Glosses}. The annotators indicated exactly when each sign began and ended (when the hand begins to fall or move into the position of another sign). Annotators did not count the anticipatory movements—while the hands get into the appropriate position to begin to articulate the sign in question—as part of that sign.  Similarly, the end of signs was identified as occurring prior to movement of the hands out of the position for that sign in preparation for the articulation of the following sign. 

The next tier was \textit{Labels for Glosses}. The annotator selected a gloss label based on an English word that represented the sign. The annotators worked for consistency in using a single correspondence English gloss for each ASL sign throughout our videos, but no controlled gloss-label vocabulary list was used for this initial gloss labeling.  However, for a specific set of 100 key glosses that were of special interest to our research project, e.g., words relating to specific grammatical structures, annotators used a controlled vocabulary of 100 gloss labels to consistently label those signs. A larger collection of video of isolated sign productions of this same set of 100 word was previously shared in a prior dataset \cite{hassan-etal-2020-isolated}. 

There was also a \textit{Fingerspelling} tier, on which annotators could identify any spans of fingerspelling in videos. There were also associated tiers to identify fingerspelling errors, e.g., use of ungrammatical handshapes, non-standard location of the hand in space, unnecessary and/or non-standard movement of the hand, and non-fluent speed of fingerspelling. The next tier group was for indicating \textit{Clauses}; annotators marked where each clause began and ended. 

There was also a set of tiers for \textit{Non-Manual Signals}. The annotators were asked to indicate specific non-manual signals (facial expressions or head movements) on the timeline. The annotator was not required to align the start-time and stop-time of each facial expression with gloss boundaries. The various types of facial expressions included: NEG (to indicate signer’s head shaking left-to-right as in a negative manner), WHQ (to indicate a WH-question facial expression), YNQ (to indicate a yes/no question), RHQ (to indicate a rhetorical question), COND (to indicate a conditional, or TOPIC (to indicate a topicalized phrase).

The final group of tiers was for \textit{Non-Manual Errors}. Annotators were asked to identify any non-manual-signal errors such as missing or incorrect facial expressions or head movements. For instance, if the signer used a negative sign like NOT or NONE but failed to produce a NEG non-manual signal. The annotators used separate tiers for errors relating to the absence of Yes-No Question, WH-Question, Rhetorical Questions, Conditional, and Negative facial expressions. 

Tier descriptions are provided in much greater detail in the ``Instruction for Using ELAN Annotation Software," which was the annotation guide and instructions document provided to annotators in this study.  This document is shared with the dataset.  

\subsection{Dataset Contents}

The dataset includes a CSV file containing demographic data for the participants, PDF files for each of the 6 homework-assignment prompts, the annotation guide and instructions document for annotators (mentioned above), original MP4 video files, Kinect v2 ``.bin" recording files, and ELAN annotation files. The ASL-Homework-RGBD dataset is available to authorized users of the Databrary platform \cite{dataset}. 


\section{Discussion, Limitations, and Future Work}

The dataset was collected to serve as training and testing data for the development of computer-vision technologies for the creation of educational-feedback tools for ASL students, i.e., systems that could analyze a video of an ASL learner and provide them feedback on their signing \cite{vahdani2021recognizing}. Beyond this initial project, we anticipate that computer-vision researchers working on designing algorithms to detect signing errors in videos of ASL can use this data to train or test their models \cite{rastgoo2021sign}. The corpus can also be used as a benchmark dataset to evaluate the performance and robustness of algorithms to detect continuous sign recognition or some specific aspects of signing, e.g. non-manual markers.

A theme of this year's workshop is how data can be made more useful for individuals beyond the field of sign-language technologies.  We anticipate that our dataset may be of interest to ASL education researchers who are investigating how the signing of ASL students compares with those of fluent signers. For instance, researchers could compare fluent and non-fluent signers across various grammatical aspects of signing, e.g., correct use of non-manual signals.  Since our dataset includes annotation of when errors occur in signing, it may also be of interest to individuals training to be ASL instructors or ASL interpreting students who wish to practice their receptive skills on non-fluent signers.

There are several limitations of this dataset:

\begin{enumerate}
   \item Each participant was not able to do all the homework assignments, leading to a variable number of homeworks and annotated videos from each participant.
   \item The data collection occurred within New York State and the participants mostly consisted of young adults. Therefore, the signs included in this dataset might not represent the wide variety of demographic and regional variation in ASL signing.
    \item We did not assess the level of proficiency of the signers. We broadly classified the signers into \textit{fluent} and \textit{non-fluent} groups. However, the actual signing fluency may be on a spectrum. 
    \item Since stimuli were shown in English, there is a risk that an ASL signer with lower English literacy may not have accurately understood the homework assignment text. To mitigate this, the sessions with fluent signers were conducted by a Deaf ASL signer researcher, who offered to clarify any details of the assignment. However, future work could consider assignment prompts based on images or other modalities. 
    \item We did not measure the level of fluency of our participants through an analysis of the videos produced or other objective measures. In future work, researchers could examine videos to assign fluency levels to participants. 
    \item The homework assignments, data collection, and annotation has been driven by the specific needs of our research project. Researchers investigating other questions may need to provide alternative or additional annotation in support of their work.
 
\end{enumerate}

\section{Acknowledgements}

This material is based upon work supported by the National Science Foundation under award numbers 2125362, 1763569, and 1462280, and Department of Health and Human Services under award number 90DPCP0002-0100. 

\section{Bibliographical References}

\begin{small}
\bibliographystyle{lrec2022-bib}
\bibliography{lrec2022}

\label{lr:ref}
\bibliographystylelanguageresource{lrec2022-bib}

\end{small}

\end{document}